\documentclass[11pt, conference]{article}
\usepackage[letterpaper, margin=1in]{geometry} 
\usepackage{graphicx}
\usepackage{amsmath}
\usepackage{amsfonts}
\usepackage{hyperref}

\usepackage{natbib}

\usepackage{url}
\makeatletter
\g@addto@macro{\UrlBreaks}{\UrlOrds}
\makeatother

\usepackage{censor}
\StopCensoring

\usepackage{tabularx}

\usepackage{hhline}

\usepackage[ruled]{algorithm2e}
\SetAlFnt{\algofont}
\SetAlCapFnt{\algofont}
\SetAlCapNameFnt{\algofont}
\SetAlCapHSkip{0pt}
\IncMargin{-\parindent}
\renewcommand{\algorithmcfname}{ALGORITHM}

\title{Artificial Intelligence for Collective Intelligence: \\A National-Scale Research Strategy}

\author{
    Seth Bullock\thanks{University of Bristol, \texttt{seth.bullock@bristol.ac.uk} (Corresponding Author)}, 
    Nirav Ajmeri\footnotemark[1], 
    Mike Batty\thanks{University College London}, 
    Michaela Black\thanks{Ulster University}, 
    John Cartlidge\footnotemark[1], \\
    Robert Challen\footnotemark[1], 
    Cangxiong Chen\thanks{University of Bath}, 
    Jing Chen\thanks{Cardiff University}, 
    Joan Condell\footnotemark[3], 
    Leon Danon\footnotemark[1], \\
    Adam Dennett\footnotemark[2], 
    Alison Heppenstall\thanks{University of Glasgow}, 
    Paul Marshall\footnotemark[1],
    Phil Morgan\footnotemark[5], 
    Aisling O'Kane\footnotemark[1], \\
    Laura G. E. Smith\footnotemark[4], 
    Theresa Smith\footnotemark[4], 
    Hywel T. P. Williams\thanks{University of Exeter}
}

\date{}

\begin{document}



%
\maketitle


\begin{abstract}

Advances in artificial intelligence (AI) have great potential to help address societal challenges that are both \emph{collective} in nature and present at \emph{national} or trans-national scale. Pressing challenges in healthcare, finance, infrastructure and sustainability, for instance, might all be productively addressed by leveraging and amplifying AI for national-scale \emph{collective intelligence}. The development and deployment of this kind of AI faces distinctive challenges, both technical and socio-technical. Here, a research strategy for mobilising inter-disciplinary research to address these challenges is detailed and some of the key issues that must be faced are outlined.

\end{abstract}

\section{Introduction}

Artificial intelligence (AI) and machine learning often address challenges that are relatively monolithic: determine the safest action for an autonomous car; translate a document from English to French; analyse a medical image to detect a cancer; answer a question about a difficult topic. These kinds of challenge are important and worthwhile targets for AI research. However, an alternative set of challenges exist that are \emph{collective} in nature:

\begin{itemize}
\item help to minimise a pandemic's impact by coordinating mitigating interventions;
\item help to manage an extreme weather event using real-time physical and social data streams;
\item help to avoid a stock market crash by managing interactions between trading agents;
\item help to guide city developers towards more sustainable coordinated city planning decisions;
\item help people with diabetes to collaboratively manage their condition while preserving privacy.
\end{itemize}

The capability of naturally occurring collective systems to solve problems of coordination, collaboration, and communication has been a long-standing inspiration for engineering \citep{swarm_intelligence_99}. However, developing AI systems for these types of problem presents unique challenges: extracting reliable and informative patterns from multiple overlapping and interacting real-time data streams; identifying and controlling for evolving community structure within the collective; determining local interventions that allow smart agents to influence collective systems in a positive way; developing privacy-preserving machine learning; advancing ethical best practice and governance; embedding novel machine learning and AI in portals, devices and tools that can be used transparently and productively by different types of user. Tackling them demands moving beyond typical AI/machine learning approaches to achieve an understanding of relevant group dynamics, collective decision-making and the emergent properties of multi-agent systems, topics more commonly studied within the growing research area of collective intelligence. Consequently, addressing these challenges requires a productive \emph{combination} of collective intelligence research and artificial intelligence research \citep{nesta2020,beritchevskaia2022}. 

In this paper we introduce and detail a research strategy for approaching this challenge that is being taken by a new national artificial intelligence research hub for the United Kingdom: \emph{AI for Collective Intelligence (AI4CI)}.\footnote{\url{https://ai4ci.ac.uk}} The AI4CI Hub is a multi-institution collaboration involving seven partner universities from across the UK's four constituent nations and over forty initial stakeholder partners from academia, government, charities and industry. It pursues applied research at the interface between the fields of AI and collective intelligence, and works to build capacity, capability and community in this area of research across the UK and beyond. 
This paper presents the AI4CI research strategy, details how it can be pursued across multiple different research themes, and summarises some of the key unifying research challenges that it must address. 

\section{Research Context}

Between 2022 and 2024, the UK government initiated several significant investments in national-scale AI research amounting to approximately £1Bn of support. Foremost amongst these investments were: the establishment of the UK's first national supercomputing facility for AI research (Isambard-AI; £225m),\footnote{\url{https://www.gov.uk/government/news/bristol-set-to-host-uks-most-powerful-supercomputer-to-turbocharge-ai-innovation}} plus an additional £500m of AI compute hardware investment across UK universities,\footnote{\url{https://www.ukri.org/opportunity/host-sites-for-the-next-wave-of-uk-government-ai-infrastructure}; N.B.\ this investment in hardware was subsequently withdrawn by the UK's incoming Labour government: \url{https://www.bbc.co.uk/news/articles/cyx5x44vnyeo}} the inception of twelve new Centres for Doctoral Training in AI (AI CDTs; £117m),\footnote{\url{https://www.ukri.org/what-we-do/how-we-work-in-ai/ukri-artificial-intelligence-centres-for-doctoral-training}} funding for a raft of AI research projects including AI for net zero (£13m)\footnote{\url{https://www.ukri.org/opportunity/artificial-intelligence-research-to-enable-uks-net-zero-target}} and AI for healthcare (£13m),\footnote{\url{https://www.ukri.org/news/13-million-for-22-ai-for-health-research-projects}} the creation of UK Responsible AI, a national network to conduct and fund research into responsible AI (UKRAI, £31m),\footnote{\url{https://www.ukri.org/news/54m-to-develop-secure-ai-that-can-help-solve-major-challenges}} and the launch of nine new national Research Hubs for AI, three focusing on the mathematical foundations of AI and six focusing on applied AI research (£100m).\footnote{\url{https://www.ukri.org/news/100m-boost-in-ai-research-will-propel-transformative-innovations}}

These significant investments were driven by growing recognition that modern AI has the potential to achieve a positive and revolutionary impact on society. Here, we focus on a research strategy proposed in response to the findings of a recent Nesta report\footnote{\url{https://www.nesta.org.uk/report/future-minds-and-machines}} 
which recommended that policymakers ``put collective intelligence at the core of all AI policy in the United Kingdom'' \citep[p.57]{nesta2020}, arguing that ``the first major funder to put £10 million into this field will make a lasting impact on the future trajectory for AI and create new opportunities for stimulating economic growth as well as more responsible and democratic AI development'' (ibid, p.58).

\section{Vision and Structure}

Our ability to address the most pressing current societal challenges (e.g., healthcare, sustainability, climate change, financial stability) increasingly depends upon the extent to which we can reliably and successfully engineer important kinds of collective intelligence, which we define as:  

\begin{quote}
``Connected communities of people, devices, data and software collaboratively sensing and interacting in real time to achieve positive outcomes at multiple scales''.
\end{quote}

Whether we are aiming to minimise the impact of a global pandemic through effectively managing successive waves of vaccination \citep{brooks-pollock2021}, to prevent financial ``flash crashes'' through effective regulation of autonomous trading agents \citep{cartlide-fast-furious-2012,johnson-etal-2013}, to make our cities sustainable and liveable through using real-time analytics to inform short-, medium- and long-term planning \citep{spooner2021,batty2024}, to combat social polarisation and climate disinformation on social media \citep{treen2020online}, or to achieve the UK NHS 2019 Long-Term Plan \citep{nhslongtermplan2019} by enabling effective healthcare ecosystems that integrate clinical care, technology, education, and social support for patients with chronic health conditions \citep{duckworth2024explainable}, invariably what is required is an ability to engineer \emph{smart collectives}.\footnote{This characterisation of collective intelligence is strongly aligned with approaches developed within socio-technical systems research \citep{baxter&sommerville2011}.}

\begin{figure}
    \centering
    \includegraphics[width=0.5\linewidth,trim={0.7cm 0.3cm 0.2cm 0.2cm},clip]{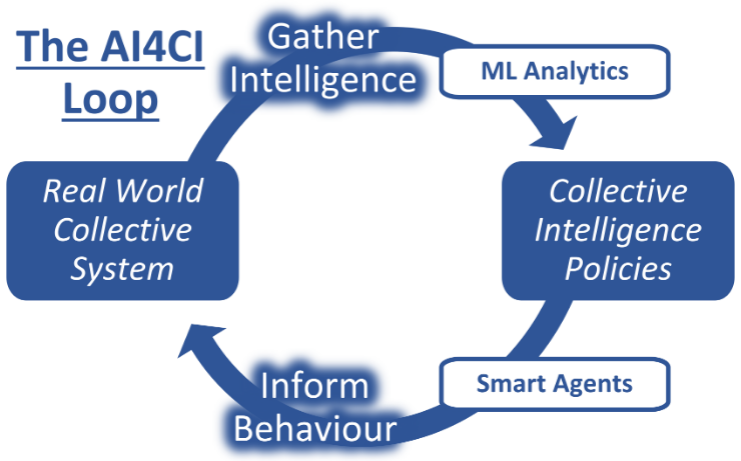}
    \hfill
    \includegraphics[width=0.475\linewidth,trim={0.1cm 0 0.2cm 0.1cm},clip]{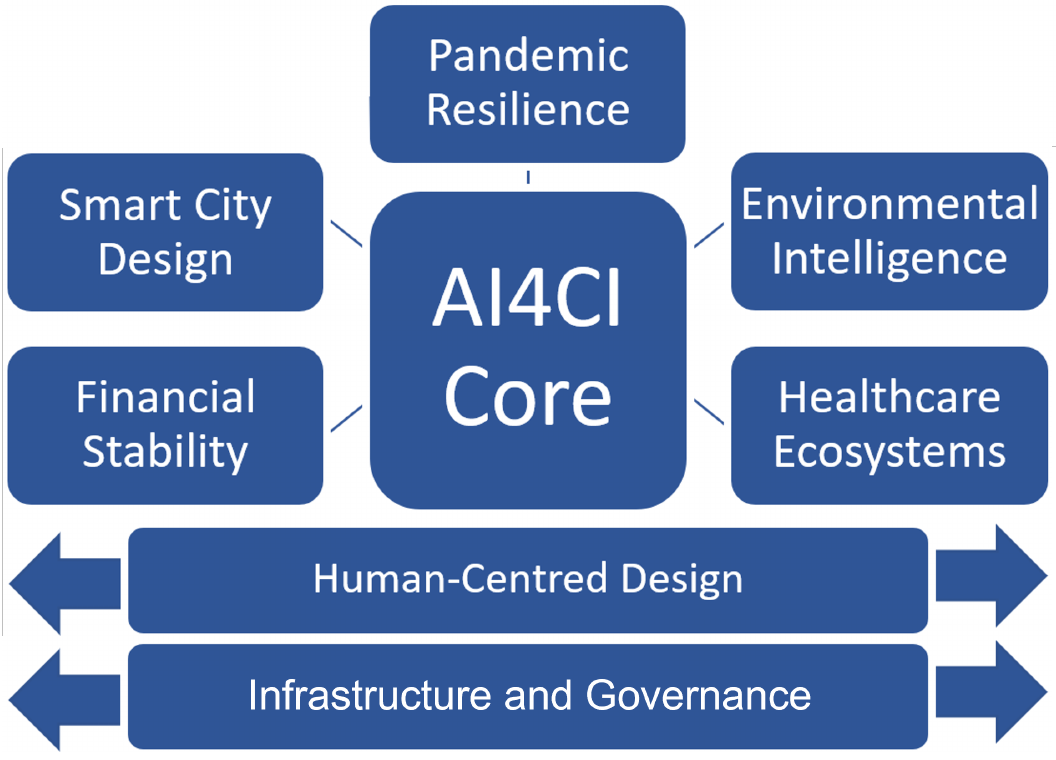}
    \caption{\emph{Left} --- The AI4CI Loop: Machine learning and AI enable distributed real-time data streams to inform effective collective action via smart agents. \emph{Right} --- The AI4CI Hub: Five applied research themes and two cross-cutting research themes are supported by the hub's central core.}
    \label{fig:loop-and-hub}
\end{figure}

This will necessarily involve addressing both halves of what we characterise as the ``AI4CI Loop'' (Fig.~\ref{fig:loop-and-hub} \emph{left}) --- (1) \textsl{Gathering Intelligence:} collecting and making sense of distributed information; (2) \textsl{Informing Behaviour:} acting on that intelligence to effectively support decision making at multiple levels. New artificial intelligence methods are unlocking progress on both halves. The first is being revolutionised by a combination of mobile devices, instrumented environments, data science, machine learning analytics and real-time visualisation, while the second is being transformed by decentralised, distributed smart agent technologies that interact directly with users. However, successfully linking both halves of the AI4CI Loop requires new human-centred design principles, new governance practices and new infrastructure appropriate for systems that deploy AI for collective intelligence at national scale.  

\section{Research Themes}
\label{sec:themes}

The AI for Collective Intelligence Hub (Fig.~\ref{fig:loop-and-hub} \emph{right}) addresses and connects both halves of the AI4CI Loop across a set of five important application domains (healthcare, finance, the environment, pandemics, and cities) and two cross-cutting themes (human-centred design and infrastructure and governance). In each domain, the challenge is to leverage and make sense of real time, dynamic data streams generated across hybrid systems of interacting people, machines and software distributed over space and across networks, in order to achieve systemic insights and drive effective interventions via the automated behaviour of smart AI agents. Pursuing research across multiple domains in concert enables each to benefit from the others' insights and maximises the chance of uncovering principles and that have domain-general application \citep{smaldino2022}.

\begin{figure}
    \centering
    \includegraphics[width=0.75\linewidth]{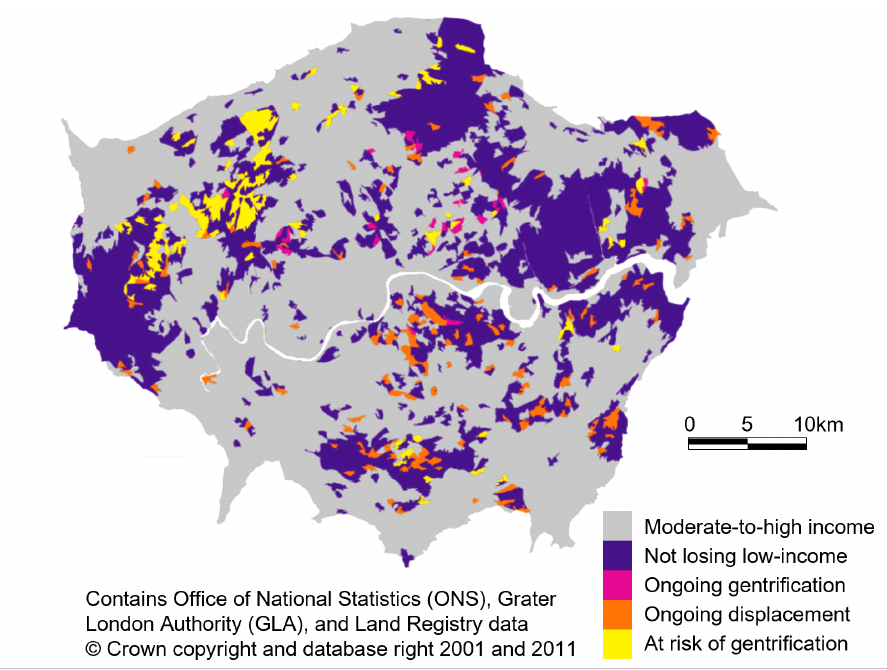}
    \caption{An indicative snapshot of smart city datasets informing AI for collective intelligence research. Gentrification and displacement typologies for Greater London in 2011 at neighbourhood level with cartogram distortion based on London's residential population in 2011. Adapted from \cite{dennett2020}.}
    \label{fig:London}
\end{figure}

\subsection{Smart City Design}

Plan-making systems for UK cities are not currently fit for purpose.\footnote{``Planning for the future'', Department for Levelling Up, Housing and Communities, UK Government, 2023, \url{https://www.gov.uk/government/consultations/planning-for-the-future/planning-for-the-future}} Local plans, the major instrument of the statutory planning system, must be modernised to exploit collective data and machine intelligence \citep{batty2024}. Meeting the challenges associated with smart planning for smart cities\footnote{The term \emph{smart city} is used here in two mutually reinforcing senses, first in the sense that novel smart technologies are physically incorporated into these cities, and second in the sense that these technologies underpin new kinds of ``smart'' behavioural interactions within and across these cities at a range of different time scales \citep{batty2020}.} in a way that delivers practical tools and applications requires integrating and exploiting multiple streams of city data provided by local and national government, urban analytics and infrastructure firms, national agencies, survey data, and human mobility patterns derived from digital traces or social media.\footnote{e.g., Data for Good: \url{https://dataforgood.facebook.com}; Smart Data Research: \url{https://www.sdruk.ukri.org} (formerly Digital Footprints); note the implicit challenges here related to (i) establishing and maintaining the public's trust in, and engagement with, these potentially intrusive data collection efforts, and (ii) countering the inevitable systematic biases that arise from unrepresentative sampling of the collective system as a whole.} 

These data can drive new AI for two purposes: (i) automating real-time intelligence for the smart city \citep{malleson2022,heppenstall2023}; (ii) informing longer-term smart city planning to meet the challenges of climate, ageing, housing affordability, and health \citep{batty2024}. Achieving smarter cities that optimise behaviours in the short and long term requires AI that extends and improves on existing models of urban structure, dealing with highly fluid situations dominated by rapid change \citep{batty2024}. This is a major challenge not only for the way that we design cities but also for how AI must deal with many/most human problem-solving contexts. Supervised and unsupervised learning methods can be used to reveal new patterns in large messy mobility datasets such as mobile phone traces, cross-validated with rich survey data to produce spatially, temporally and attribute rich insights into the seismic shift in post-COVID mobility patterns \citep{batty2020}. Predictive tools for the design of new patterns of transport and land development at different scales can be founded on models that take multiple land suitability and mobility indices as inputs \citep[see, e.g., Fig.~\ref{fig:London};][]{dennett2020}. Deriving meaningful interpretations of these models and enabling decision-makers to explore how optimal plans play out over time and space in the context of synthetic AI agent models \citep{heppenstall2012} delivers the explanatory accounts that are essential for public accountability in the use of these methods for city decision making.

\subsection{Pandemic Resilience}

\begin{figure}
    \centering
    \includegraphics[width=0.48\linewidth]{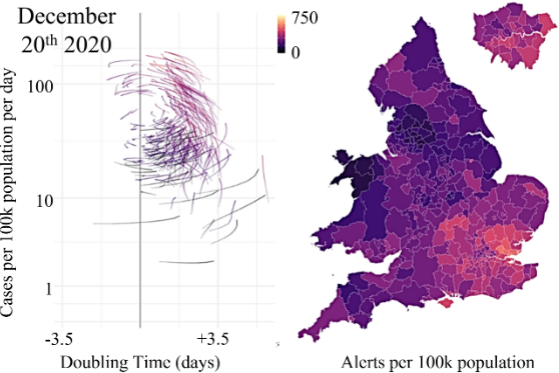}
    \hfill
    \includegraphics[width=0.48\linewidth]{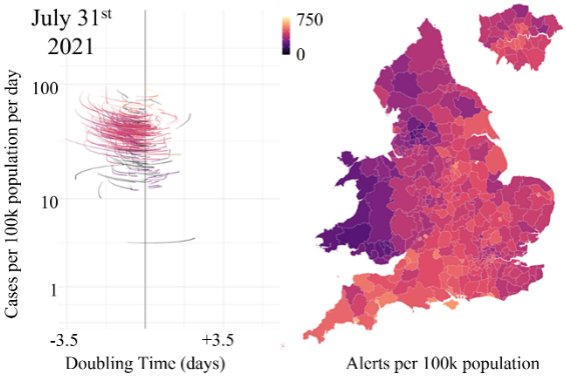}
    \caption{A snapshot of pandemic datasets informing AI for collective intelligence research. Regionally disaggregated datasets relate the level and growth rate of COVID-19 cases (phase plots) with the rate of digital contact tracing alerts delivered to citizens by the NHS mobile phone app (maps) at two points in time during the COVID-19 pandemic. \emph{Left} --- December 20$^{\mathrm{th}}$ 2020: the alpha variant is spreading in the south-east despite a `circuit-breaker' lockdown. \emph{Right} --- July 31$^{\mathrm{st}}$ 2021: Digital contact tracing alerts are triggered by high COVID-19 case burden}
    \label{fig:pingdemic}
\end{figure}

COVID-19 exposed weaknesses in the UK's pandemic resilience. A combination of collective intelligence and AI can help us do better next time. Data crucial for managing novel pandemics are inherently fragmented, arising from communities of medics, public health professionals and analysts to describe the spread of the disease, characterise its phenotype, and, with appropriate modelling, inform appropriate policies nationally and locally \citep{brooks-pollock2021}. National spatio-temporal datasets describing SARS-CoV-2 hospital testing and community testing, and the extent and effects of the mitigations put in place against it, can be exploited in order to build new AI/machine learning tools for future pandemics---and potentially for mitigating seasonal outbreaks of endemic disease. 

Two strands of research can be identified. First, a suite of machine learning models fuelled by national SARS-CoV-2 pandemic data can be used to explore and demonstrate how the integration of multiple population-level indicators could have improved decision making during the pandemic. Due to the urgency of the need for response during the pandemic, developing and validating this kind of analytical infrastructure was not possible. With it in place, however, challenges associated with imperfect data can be addressed. For example, data from unreliable laboratories distorted local and national COVID-19 reproduction number ($R_t$) estimates in 2022\footnote{\url{https://www.gov.uk/government/news/ukhsa-publishes-investigation-findings-following-errors-at-the-private-immensa-lab}} leading to under-informed policy decisions. Automatic detection and correction for such errors will alert policy makers and decision makers more quickly. 

Second, detailed local data can be used to understand localised interventions and spontaneous behavioural responses. Here, relevant AI challenges include: establishing the relative value of diverse data streams at different stages of the pandemic and their consistency across spatial scales; the use of anomaly and change point detection to identify meaningful discontinuities; robust data imputation, pattern completion, bias detection and correction; evaluating the impact of both vaccination and behavioural change resulting from, for example, non-pharmaceutical interventions (and their interactions); and coping with delay and bias in data capture and clinical outcome results during the exponential growth phase of a new variant. One particular focus is the impact of contact tracing apps on the behaviour of individuals and the public-health messaging around their use (see Fig.~\ref{fig:pingdemic}).

Two major challenges cut across both strands: quality assurance within a privacy protecting framework \citep{Challen2019} and managing the regional heterogeneity of pandemic impact and associated behavioural change \citep{Challen2020}. 

If these can be overcome, there is the potential to deliver a suite of tools (sensitive to local population properties: income, mobility, demographics) that inform policy on where and when to increase or decrease testing capacity, implement contact tracing and/or isolation measures, distribute limited hospital capacity, etc. By working in collaboration with key stakeholders in government, a set of interactive portals can be developed that effectively inform policy decisions (by reporting well understood metrics: the reproduction number, $R_t$, hospitalisations, excess deaths) and/or individual behaviour (e.g., by presenting bespoke risk scenarios to increase adherence to government-imposed restrictions).

\subsection{Environmental Intelligence}

In order to meet the challenge of mitigating climate change, there is a need to improve access to, and comprehension of, different kinds of complex, time-varying environmental data, for example: (1) outputs from climate, weather and ocean model ensembles and empirical observations, (2) high-volume geospatial, ecological, satellite and remote sensing data; (3) socioeconomic data on resource flows, supply chains, energy consumption and carbon emissions, and (4) online media including news media and cross-platform social media content. Many decision-makers (including citizens and policy-makers) would benefit from better environmental information. A huge volume of this data is now available, but it often requires a high level of expertise to obtain it and interpret the associated uncertainties. Large language models are increasingly suggested as a potential solution to part of this problem \citep{vaghefi2023chatclimate,koldunov2024local}. Meanwhile, public debate is weakened by the profusion of poor quality or deliberately false information, especially concerning the contested issue of climate change \citep{treen2020online,acar2023}.  A combination of AI and collective intelligence approaches can deliver tools that overcome these challenges by democratising access to good quality information about environmental change.

Novel ``climate avatar'' agents can act as simple interfaces between complex environmental data and the people who need it in order to improve their decision-making.  They ingest weather and climate data from existing large datasets,\footnote{e.g., \url{https://www.metoffice.gov.uk/research/approach/collaboration/ukcp}, the UK Met Office's Climate Projections dataset.} peer-reviewed climate science literature and other trusted sources (e.g., IPCC reports\footnote{\url{https://www.ipcc.ch/reports/}}), and expose this data via natural-language interfaces that allow users to access information and gain understanding in a conversational style. By summarising scientific literature and generating on-the-fly visualisations from raw data, climate avatars enable lay users to make sense of complex climate data and uncertainties, tailored to their specific context (e.g., where they live or the sector in which they work). Similar smart agents engage with other environmental data sources, such as geospatial data, satellite imagery and ecological data. Once created, validated and trusted, these avatars can be deployed to interact with human users in different contexts, for example: allowing expert and non-expert academics to interrogate complex federated models of natural and human capital; enabling chatbots to explain extreme weather events and provide warnings/guidance; providing timely responses to policy formulation queries; or defusing toxic discourse on social media \citep{treen2020online}. Data ethics, governance and usability challenges must be addressed in order to ensure that agents are explainable, trustworthy, and able to effectively influence public understanding in order to achieve positive social outcomes.


\subsection{Financial Stability}

Modern financial technology (FinTech) presents major challenges for both regulation (cf. the UK Government's 2021 Kalifa Review\footnote{\url{https://www.gov.uk/government/publications/the-kalifa-review-of-uk-fintech}}) and consumer protection/trust (cf. the UK Financial Conduct Authority's 2022 Consumer Duty\footnote{\url{https://www.fca.org.uk/firms/consumer-duty}}). These challenges can be addressed through collaboration with relevant SMEs, non-profits, consultancies, national research organisations, data providers, platform providers and national FinTech hubs, to co-create personalised AI for early warning indicators, informing financial decision making, and reducing vulnerability to manipulation.

The development of personalised, adaptive AI driven by collective intelligence derived from financial systems to enable individuals, businesses and government to make better-informed decisions can be considered in two contexts: (i) financial markets \citep{shi-lob-prediction-22,maggie-mifid-23,liu2023,zhang2023,shi-agent-LOB-24,you-stock-prediction-24} and (ii) personal finances \citep{imf-financial-inclusion-credit,vanthiel2024}. Progress on either requires that systems take advantage of a range of rich and often non-traditional financial datasets: e.g., high frequency financial trading data, equity investment data for retail traders, personal finance, lending records and psychometric credit ratings, demographic household data, NLP-enhanced social media and news, and FCA-approved synthetic datasets used for testing all aspects of financial technology for the UK's Financial Conduct Authority regulatory sandbox.

Models that leverage social media, news media and traditional data (prices, volumes, etc.) can provide regulators with early warning signals of bubbles/crashes and protect non-professional investors from pathological investment behaviour \citep[e.g., the viral herding that led to the recent GameStop short squeeze;][]{klein-gamestop-2022,kudzanai2023} with model validation taking place within the Financial Conduct Authority sandbox.\footnote{\url{https://www.fca.org.uk/firms/innovation/digital-sandbox}} AI assistants that enable collective ethical investment, and protect vulnerable households from exploitative personal finance providers can be co-created with relevant charities, e.g., working with data-driven psychometric credit rating tools that personalise the immediate and longer-term implications of personal and SME borrowing decisions. Overall, work in this area can make use of rich forms of non-traditional financial data to develop new AI tools that can de-risk the rapid ``democratisation of finance'' that is being provoked by the ongoing FinTech revolution \citep{arner-etal-2015}.

\subsection{Healthcare Ecosystems}


The ability of the UK's NHS to deliver its Long-Term Plan depends critically on its capacity to automate bespoke monitoring and support for a range of long-term health conditions at population scale \citep{Topol_2019}. This cannot be achieved without a step change in the use of longitudinal data analysis and smart software assistants. Doing so will require working in close collaboration with clinical partners and technology firms on healthcare analytics and the design of user experience for healthcare AI.

Anonymised patient records that track, e.g., mental health consultations or diabetes progression are complex, partial and noisy reflections of longitudinal patient trajectories with potential to improve clinical decision making and empower patients \citep{rajkomar2019machine}. The UK's NHS trusts and healthcare technology firms have extensive expertise in leveraging data to manage and treat these conditions, including cohorts of diabetes patients engaged in the co-design of AI systems for collective intelligence that are trustworthy and effective \citep{duckworth2024explainable}.

Here, two interacting strands of research can be identified: (i) machine learning analytics for collective healthcare data, and (ii) co-design of smart healthcare agents for patient collectives. Strand (i) develops methods for unsupervised extraction and quantification of patterns from patient data pooled across heterogeneous sources from clinical systems to networked personal devices in order to discover clusters in symptom trajectories, detect adverse events, and recommend treatment and self-care strategies. This work leverages cutting-edge privacy preserving and federated machine learning methodologies to enable machine learning on data across all the sources interactively and in real-time while guaranteeing that the identity of the individual patients and the data they provide will not be leaked through for example training-data leakage attacks \citep{ChenAnalysingTrainingDataLeakage2022}. Strand (ii) works with hard-to-reach patients (those suffering from secondary health conditions, mental health conditions, or living circumstances that prevent them from accessing health care unaided and limit their use of technology), plus their carers and clinicians, to address issues of trust, usability, and efficacy in ethical AI for informing healthcare decision making across patient populations \citep{stawarz}. The over-arching challenge for both strands is to leverage population-wide data collection for informing robust individualised decision-making without compromising anonymity and under realistic data and user assumptions. A key challenge is using AI to mediate between patients and care systems rather than burdening already overloaded clinicians with another software tool \citep{emanuel2019artificial}.  

\section{Cross-cutting Themes}

The application domains described above are by no means the only areas in which AI for collective intelligence has strong potential. Additional problems for which productive work could have significant transformative effects include preventing violent extremism \citep{smith2020,alife2023a}, addressing the climate crisis \citep{santos2019}, collaborating with autonomous systems \citep{swarmintel2018,hart2022}, and reducing energy consumption \citep{pitt2018}. However, in addition to confronting issues specific to each of these individual use cases, achieving AI for collective intelligence also faces challenges that cut across these application areas.

\subsection{Human-Centred Design}

One issue vital to developing AI for collective intelligence within any use domain is achieving successful interaction with human users. Methods from social and cognitive psychology and human factors must be integrated with the various kinds of research activity outlined above in order to derive human-centred design principles for effective, trustworthy AI agents that inform behavioural change at scale within socio-technical human-AI collectives. 

Three parallel strands can be identified: (i) bringing human-centred design considerations to the work within various domain-specific AI for collective intelligence research themes; (ii) developing usable smart agents that assist users in accessing, understanding, and acting on guidance derived from collective intelligence data; and (iii) pursuing fundamental questions related to understanding and managing ``tipping points'' in collective intelligence systems. For (i), participatory design methods \citep{bratteteig2012} involving academics and stakeholders can be employed to prototype human-machine interfaces (HMIs) for the AI systems being developed. For (ii), data from human experiments can inform a series of design iterations, focussing on accessibility, usability, explainability, adaptability and trust \citep{choung2023}, drawing upon long-standing approaches to defining and measuring trust in automation \cite[e.g., ][]{lee2004}, with all being key factors for the acceptance, adoption and continued use of new technologies. Comparative analyses of these data reveal transfer effects between different theme settings, guiding development of demonstrators within each domain. For (iii), testable predictions of how to identify, characterise and influence tipping point thresholds for behaviour change can be derived from data on explainability, confidence, persistent adoption, praise and blame, e.g., based on the perceived capability of the system \citep{zhang2024}. 

Rigorous empirical methods (including controlled experiments and human simulations) must be informed by relevant psychological theory \citep[e.g., Gibsonian affordances; see, e.g.,][]{greeno1994}, human factors approaches \citep[e.g., hierarchical task analysis; see, e.g.,][]{stanton2006}, tools \citep[e.g., vigilance protocols,][]{al-shargie2019} and measures \citep[e.g., of situational awareness and cognitive load;][]{situationalawareness,cognitiveload}, and tipping point analytics \citep[e.g., autocorrelation and critical slowing down measures,][]{scheffer2009}. Frameworks for technology acceptance and adoption \citep[e.g., ``designing for appropriate resilience and responsivity'',][]{chou2023} can be employed to measure trust in new technology. Of equal importance is being able to optimally measure loss of trust (which can and will happen, e.g., due a negative experience) and crucially how to restore it---all of which will likely involve ensuring that human-centred design from prototype to deployment considers factors including system accessibility, functionality, usability and adaptability.

Combining the three strands ensures that research in each domain translates into usable, trustworthy demonstrator systems supported by insights into smart agent adoption, trust and trust restoration, and that methods for anticipating and influencing collective change inform interaction design principles for practitioners developing and employing AI systems for collective intelligence across multiple socio-technical domains.

\subsection{Infrastructure and Governance}

A second cross-cutting issue vital to deploying AI for collective intelligence at scale in any use domain is ensuring that such systems have robust infrastructure and governance (I\&G) guided by appropriate regulations and principles, delivering trustworthy AI systems and solutions that are human-centred, fair, transparent, and interpretable for the diverse range of end users.

New I\&G tools and guidelines for national scale collective AI systems that ensure privacy, quality and integrity of data, and control access to data and systems across relevant AI infrastructures \citep{aarestrup2020,shi-midas,cao2023} should be informed by experiences with previous large-scale AI platforms, such as that developed within the EU-wide MIDAS project \citep{black-midas}. One key context in which to pursue this challenge is work developing effective applied AI for national-scale healthcare systems \citep[e.g., cancer, dementia, arthritis;][]{condell-cancer,condell-dementia,condell-arthritis}. The data sensitivity and outcome criticality of these challenges for AI makes this an ideal domain in which to develop effective I\&G tools and thinking. However, considering these issues across multiple diverse applications domains also enables the unique and novel aspects of those settings to inform new thinking on I\&G questions. 

Such research must address ethical, legal, and social aspects (ELSA) of I\&G \citep{van2021elsa}, embedding ELSA accountability in robust governance frameworks and data infrastructure plans. For instance, studies should embed ELSA, FAIR Principles\footnote{\url{https://www.go-fair.org/fair-principles/}} and the Assessment List for Trustworthy Artificial Intelligence (ALTAI)\footnote{\url{https://digital-strategy.ec.europa.eu/en/library/assessment-list-trustworthy-artificial-intelligence-altai-self-assessment}} into their Data Management Plans (DMPs) and infrastructure governance and should be informed by outputs of the Artificial Intelligence Safety Institute (AISI)\footnote{\url{https://www.gov.uk/government/publications/ai-safety-institute-overview/introducing-the-ai-safety-institute}} and other relevant guidelines, e.g., the EU's Ethics Guidelines for Trustworthy AI,\footnote{\url{https://digital-strategy.ec.europa.eu/en/library/ethics-guidelines-trustworthy-ai}} and relevant regulation frameworks such as the EU AI Act.\footnote{\url{https://eur-lex.europa.eu/legal-content/EN/TXT/?uri=CELEX:52021PC0206}} 

Finally, the prospect of truly national-scale AI systems of the kinds being considered here foregrounds the pressing need for truly \emph{trans-national} governance structures and mechanisms. These are particularly relevant in collective intelligence settings, since the people, diseases, finance, etc., at the heart of such systems, and the data pertaining to them, all transcend national boundaries. As the Final Report of the United Nation's AI Advisory Body puts it ``the technology is borderless'', necessitating the establishment of ``a new social contract for AI that ensures global buy-in for a governance regime that protects and
empowers us all'' \citep{un2024}.

\section{Research Strategy}

To make significant progress across the research strands outlined above, an effective AI for collective intelligence research strategy must also consider the set of meta-level research challenges that must be overcome if academic research findings are to translate into effective and impactful real-world outcomes. There include addressing underpinning issues around stakeholder engagement; equality, diversity and inclusion (EDI); environmental sustainability; and responsible research and innovation (RRI).

\subsection{Stakeholder Engagement}

We distinguish here between three categories of research stakeholder relevant to AI for collective intelligence research: data partners, skills partners and academic partners. These categories are not disjoint since a single organisation may play more than one of these roles, but they do serve to distinguish between different kinds of research interaction that may be necessary in order to achieve successful applied research in the AI for collective intelligence space at national or trans-national scale.

\emph{Data Partners} are ``problem owning'' organisations willing to provide controlled access to data, expertise, tools, personnel and strategic guidance relevant to a societal challenge or user need that can be addressed by AI for collective intelligence research. For national-scale efforts, these will tend to be national or trans-national agencies (e.g. the UK's National Health Service or the UK Health Security Agency) and departments within national government (e.g., the UK's Department for Health and Social Care), but may also include commercial outfits such as pharmaceutical firms involved in vaccine development, etc. Crucial issues for research collaboration here include those surrounding intellectual property, privacy, regulatory frameworks (e.g., GDPR in the EU), secure data hosting, etc. There are also challenges around the emerging role of synthetic data as a substitute for data that is too sensitive to share or is too hard to anonymize. Such synthetic data can be useful where a mature understanding of the underlying real-world system and the data generating process is in place, but can be problematic in the absence of such an understanding since it can be difficult to provide assurances that the synthetic data captures all of the necessary structural relationships that are present in the original (poorly understood) dataset \citep{10.1145/3630106.3659002}. More generally, issues around incomplete or noisy data or data that is not sufficiently representative of the underlying population are familiar problems that have significance here. 

\emph{Skills Partners} are ``problem solving'' organisations that are already involved in pioneering the AI and collective intelligence skills, tools and technologies that are driving the next generation of AI for collective intelligence applications, e.g., multi-agent systems, collective systems data science, advanced modelling and machine learning, AI governance and ethics, etc. These may include blue-chip outfits and national facilities (e.g., the UK's Office of National Statistics) but will also include many of the small and medium-sized enterprises (SMEs) emerging in this space (e.g., Flowminder,\footnote{\url{https://www.flowminder.org/}} who leverage decentralised mobility data to support humanitarian interventions in real-time). Research opportunities here include connecting innovating skills partners to the data partners that require their expertise while navigating the intellectual property and commercial sensitivity issues that surround an emerging (and therefore somewhat contested)  part of the growing AI consultancy sector. 

\emph{Academic Partners} are individuals, research groups or larger academic research activities that are operating in the AI for collective intelligence space. This is a growing area of activity and a key challenge here is connecting and consolidating the emerging community and linking it effectively with the two categories of non-academic stakeholder described above. One key challenge for academic research in this space is balancing the need for rigorous well-understood and mature theory and methods in order to provide quality assurances and guarantee robustness of AI system behaviour against the need to explore and develop new and improved theory and methods that take us beyond the current limited capabilities of extant tools and systems.

\subsection{Equality, Diversity and Inclusion}

The AI workforce lacks diversity \citep[e.g.,][]{young2021}. Moreover, AI technology can tend to impose and perpetuate societal biases \citep[e.g.,][]{kotek2023}. Consequently, it is important that AI for collective intelligence research \emph{operations} be a beacon for best practice in equality, diversity and inclusion (EDI). Moreover, an effective AI for collective intelligence research strategy should itself also be driven by equality, diversity and inclusion \emph{research considerations}. The emerging ``AI Divide'' separating those that have access to, and command of, powerful new AI technologies from those that do not threatens to further marginalise under-represented, vulnerable, and oppressed individuals and communities \citep{wang2024}. Collective intelligence methods sometimes focus on achieving consensus and collective agreement (which can tend to prioritise majority views and experiences). However, in addition to aggregating signals at a population level in order to inform the high-level policies and operations of national agencies (which will themselves benefit from being sensitive to population heterogeneity), the AI for collective intelligence research described here is equally interested in deriving bespoke guidance for individuals (or groups) that respects their specific circumstances and needs. This aspect of the research strategy explicitly foregrounds the challenge of reaching and supporting diverse users and those that are intersectionally disadvantaged, e.g., diabetes patients that also have mental health conditions \citep{price2023}.

\subsection{Environmental Sustainability}


The carbon footprint of most academic research is dominated by travel \citep{Achten2013}. Consequently, research in this area, like any other, should seek to minimise the use of flights and consider virtual or hybrid meetings wherever possible. Other steps that can be taken to reduce the environmental impact of research practice and move towards ``net zero'' and ``nature positive'' ways of working include making sustainable choices for procurement (e.g., accredited sustainable options) and catering (e.g., plant-based food choices).

Moreover, the environment can itself be the focus of AI for collective intelligence research (see ``Environmental Intelligence'', above) or a key driving factor (see ``Smart City Design'', above). Many associated research themes are consistent with a sustainability agenda in their motivation to achieve effective interventions at scale without consuming vast resources. Intended outcomes and technologies aim to transition society to more sustainable practices (e.g., by using healthcare resource more eﬀiciently, by encouraging sustainable cities, etc.). 

However, in common with AI research more generally, AI for collective intelligence makes use of energy-intensive technologies. Computational research is energy-intensive: machine learning incurs high CPU/GPU energy cost for training \citep{patterson2021}, while storage/transfer/duplication of large datasets consumes energy in data centres. Hardware components use rare metals linked to environmental damage and inequalities. Consequently, the environmental impacts of AI for collective intelligence research should always be considered and reduced using, inter alia, computational resources powered by renewable energy, energy efficient algorithms and coding practices, and minimal data duplication. The technologies developed through this kind of research should be evaluated using full Life Cycle Assessment (LCA) techniques that measure their direct impacts (e.g., production, use and disposal costs), indirect impacts (e.g., rebound effects that increase carbon emissions elsewhere in the economy), and identify possible mitigations (e.g., substitution and optimisation effects) \citep{schien2019}.


\subsection{Responsible Research and Innovation}

Researchers can never know with certainty what future their work will produce, but they can agree on what kind of future they are aiming to bring about, and work inclusively towards making that happen \citep{owen2013,stilgoe2020}. For AI for collective intelligence research, this means working with diverse end users and stakeholders to produce a future in which national-scale AI for collective intelligence systems are tools for societal good \citep{leonard2022}. 

There are several reasons for taking responsible research and innovation (RRI) concerns especially seriously in the context of artificial intelligence research projects. First, since AI is one of the 17 sensitive research areas named in the UK's National Security and Investment Act,\footnote{\url{https://www.gov.uk/government/publications/national-security-and-investment-act-guidance-on-notifiable-acquisitions}} particular care must be taken by UK universities when establishing and pursuing AI research collaborations. Stakeholder partners must be vetted, and input from the UK Government's Research Collaboration Advisory Team (RCAT)\footnote{\url{https://www.gov.uk/government/organisations/research-collaboration-advice-team}} must be sought where there are concerns regarding, for instance, the exploitation of intellectual property arising from the research activity. Moreover, applied AI research is often fuelled by data that is sensitive, meaning that huge care must be taken to safeguard this data and ensure privacy through the use of, e.g., secure research data repositories that feature robust controlled data access protocols. More generally, since AI innovations have the potential to radically reshape the future in ways that are very hard to predict, articulating a clear shared vision of the future that is being aimed for is particularly important. Finally, AI researchers have a responsibility to engage with the public discourse around AI which is currently driving considerable anxiety and confusion.\footnote{\url{https://www.ey.com/en_us/consulting/businesses-can-stop-rising-ai-use-from-fueling-anxiety}}

\section{Unifying Research Challenges}

The research strategy outlined here sets out to develop, build and evaluate systems that exploit machine learning and AI to achieve improved collective intelligence at multiple scales: driving improved policy and operations at the level of national agencies and offering bespoke guidance and decision support to individual citizens. 

Why are systems of this kind not already in routine operation? Many of the component technologies are established and some are becoming reasonably well understood: recommender systems \citep{resnick1997}, machine learning at scale \citep{Lwakatare2020}, networked infrastructure and Internet devices \citep{radanliev2020,rashid2023}, conversational AI \citep{kulkarni2019conversational}, network science analyses \citep{networkscience}, etc. However, several important and interacting challenges obstruct the realisation of AI for collective intelligence and these must be targets for this research effort. Here, we distinguish three categories: \emph{human} challenges, \emph{technical} challenges and \emph{scale} challenges. Each application domain manifests a combination of challenges in a distinctive way (see Table~\ref{tab:challenges}), but since these challenges are inter-related they should be approached holistically.\footnote{One theoretical framework with promising potential to support and inter-relate the challenges being considered here is that offered by studies in cumulative cultural evolution \citep{mesoudi2018,smaldino2014}.}

\begin{table}
    \caption{Examples of how three different categories of unifying research challenge apply within five different AI for collective intelligence application domains.}
    \label{tab:challenges}
    \centering
    \small
    \begin{tabularx}{\textwidth}{ >{\raggedright\arraybackslash\hsize=.64\hsize}X | >{\raggedright\arraybackslash\hsize=1.12\hsize}X >{\raggedright\arraybackslash\hsize=1.12\hsize}X >{\raggedright\arraybackslash\hsize=1.12\hsize}X }
      Application\newline Domain         & Human Challenges & Technical Challenges & Scale Challenges \\
      \hhline{ =|= = =}
      
      Smart City\newline Design          & Delivering informative visualization of complex city dynamics 
                                         & Predicting outcomes of interventions that trigger behavioural change 
                                         & Balancing short-term optimisation and medium term sustainability \\
      \hline
      Pandemic\newline Resilience        & Achieving and maintaining trust in guidance/interventions 
                                         & Predicting reproductive number under phased vaccination efforts 
                                         & Differentiating regional and national collective intelligence\\
      \hline
      Environmental\newline Intelligence & Providing transparent provenance indications for complex climate data 
                                         & Evaluating climate avatar performance in the wild
                                         & Coping with bursty demand during extreme weather events \\
      \hline
      Financial\newline Stability        & Incorporating user values in personal finance recommendations 
                                         & Achieving privacy-\newline preserving ML for financial analytics
                                         & Operating at speeds consistent with flash crash cascade dynamics \\
      \hline
      Healthcare\newline Ecosystems      & Achieving effective usability for hard to reach patients 
                                         & Identifying continuously evolving clusters of similar patient types 
                                         & Achieving persistent uptake at scale for patients and NHS trusts \\
      \hline
    \end{tabularx}
\end{table}

\subsection{Human Challenges}

In order to be successful, the systems developed at the boundary between AI and collective intelligence must actually be engaged with and used by individual people as well as by institutions and agencies. For this to be the case, these systems must be trusted. Individual people must trust the systems with their data and must trust the guidance that they are offered.\footnote{There is some debate as to whether notions of `trust' and `trustworthiness' are appropriate for framing the legitimacy of AI systems; compare, for example, the work of \citet{andras2018} with the position of \citet{bryson2018}. Even the use of a term like `guidance' to describe the kind of support that AI systems might be designed to offer can be reminiscent of previous attempts to shift public behaviour through the `libertarian paternalism' of nudge economics, an approach that was discredited precisely because of its tendency to disempower or even coerce people rather than fully inform or partner with them \citep{goodwin2012}.} Institutions and agencies must trust the reliability of the aggregated findings delivered by the systems and must trust that the systems will operate in a way that does not expose them to reputational risk by disadvantaging users or putting them at risk. The European Commission's High-Level Expert Group on Artificial Intelligence suggests that trust in AI should arise from seven properties: empowering human agency, security, privacy, transparency, fairness, value alignment, and accountability.\footnote{\url{https://digital-strategy.ec.europa.eu/en/library/ethics-guidelines-trustworthy-ai}} 

What are the hallmarks of these properties that users of AI for collective intelligence systems intuitively and readily recognise? What kinds of guarantees for these properties could be credibly offered to regulators or law makers? More generally, how can users of all kinds become confident that these properties are present in a particular system and remain confident as they continue to interact with it?

Moreover, since the systems envisioned here purport to offer \emph{bespoke} decision support tailored to the needs of individual users, one acute aspect of this challenge relates to supporting the needs of all kinds of user including those from marginalised or under-represented groups. It is typically the case that machine learning extracts patterns that generalise over the diversity in a data set in order to capture central tendencies, robust trends, etc. This can fail to capture, respect or represent the features of dataset outliers. Within society, these outliers are often individuals particularly in need of support, and this support may not be useful unless it is sensitive to the specific features of these individuals' circumstances. Meeting this challenge, and the more general challenge of deserving, achieving and maintaining trust, will require an interdisciplinary combination of both social and technical insights \citep{eps364670}.

\subsection{Technical Challenges}

Amongst the many technical challenges that must be overcome to enable AI for collective intelligence to be effective, we will mention three: nonstationarity in collective systems, privacy and robustness of multi-level machine learning, and the ethics of multi-agent collective decision support.

All machine learning makes a gamble that the future will resemble the past, yet we know that the data from collective systems can be nonstationary, i.e., these systems can make transitions between regimes that may differ radically from one another \citep{scheffer2009}. How can we anticipate and detect these sudden shifts, phase transitions, regime changes, and tipping points at the level of entire collectives, sub-groups and even individuals? These questions have been considered within collective intelligence research \citep{mann2022,tilman2023}, and the large scale of systems under study here offers potential to trial methods that have been applied to physical and biological systems, e.g., early warning signals from dynamical systems theory \citep{scheffer2009}, but can these be effective for complex fast-moving socio-technical systems?

Amongst the many other machine learning challenges relevant here, we will highlight two that arise as a consequence of the fundamentally multi-level nature of collective intelligence. The AI4CI Loop (Fig.~\ref{fig:loop-and-hub}) depicts the way in which the approach to AI for collective intelligence being pursued here involves machine learning models that deliver findings at different levels of description, from findings that characterise the entire collective through intermediate results related to sub-groups within the collective to bespoke results relevant to individual members of the collective. Delivering this requires (likely unsupervised) methods to cluster and/or unbundle heterogeneous data stream trajectories derived from groups and individuals. 
Moreover, achieving this whilst maintaining the privacy of individual members of the collective requires robust privacy-preserving machine learning methods. Employing foundation models to capture and compress the patterns in the collective system is one possible approach, but understanding the vulnerabilities of these models remains an open research challenge \citep{chen2023understanding,messeri2024}.

Within collectives, one member's actions can affect other members. In this context, systems that support decision making do not only impact their direct user \citep{Ajmeri-IJCAI18-Poros,vinitsky2023}. While there is potential to leverage this collective decision making to achieve efficient coordinated outcomes \citep{eps267064}, it remains the case that a typical AI agent tends to cater to the interests of their primary user even if they are intended to reflect the preferences of multiple stakeholders \citep{Murukannaiah-AAMAS20-BlueSky}. This may reinforce existing privileges and could worsen the challenges faced by vulnerable individuals and marginalised groups. Thus, it is imperative that AI agents consider and communicate the broader collective implications of the decision support that they offer. In this way we can encourage these agents to respect societal norms and their stakeholders' needs and value preferences, and inform decisions that promote fairness, inclusivity, sustainability and equitability \citep{Murukannaiah-AAMAS20-BlueSky,Woodgate+Ajmeri-AAMAS22-BlueSky,Woodgate+Ajmeri-CSUR2024-Principles}.

\subsection{Scale Challenges}

This paper has articulated a set of research challenges in terms of ``producing national-scale AI for collective intelligence''. For some domains, e.g., pandemic response, this scale might appear to be a natural level of description because relevant policy, operations, data and governance are all ultimately defined at the level of national government. However, most if not all collective intelligence challenges engage with multiple spatial, social and governmental scales. Decision making at national, regional, local, household and personal scales are simultaneously in play, and in some cases trans-national scales are also significant as when pandemics, environmental disasters or financial contagion cross national borders. Consequently, the adjective ``national-scale'' should not be taken here to imply a single scale of operation and a single locus of decision making. Rather, the most effective AI for collective intelligence systems will be able to operate in a hierarchical, cross-scale fashion as anticipated in the work of, e.g., \citet{ostrom2010} and others.

Operating at national or trans-national scale can help to address some of the human and technical challenges discussed in the previous sections: national agencies, such as the UK's NHS or Met Office, are often trusted agencies; engaging with large populations of users can enable better support for marginalised groups of users that are typically under-represented; operating at scale can increase sensitivity to nonstationarity in an underlying collective system. However, scale also brings its own challenges in terms of establishing and managing appropriate infrastructure in a way that is secure, well-governed and sustainable.

Such infrastructure must support data collection at massive volume. Services must be delivered at point of use, in real time, without failure. Sensitive personal data must be handled securely and systems must respect the privacy of individuals while also providing solutions that rely on data aggregation and sharing. Since the security of national digital infrastructure is increasingly important in a global context where cyberattacks and information operations are becoming more common, infrastructure and service delivery must be robust to external threats as well as internal perturbations and flaws. The infrastructure and associated services and processes must be subject to appropriate and effective governance. Finally, environmental sustainability must be a core aim. How can the national scale systems and services envisioned here operate in a way that has low environmental impact?

\section{Conclusion}

There is considerable potential for productive research at the intersection between the fields of artificial intelligence and collective intelligence. This paper has presented one research strategy for operating at this intersection, articulated in terms of the AI4CI Hub's research vision and research themes, its approach to prosecuting interdisciplinary, collaborative research, and its set of unifying research challenges. The Hub's first steps include pursuing case study research projects within each of the AI4CI themes in collaboration with relevant stakeholders,\footnote{\url{https://ai4ci.ac.uk/research/}} hosting workshops and symposia to cross-fertilise and disseminate new tools, methods and thinking at the AI/collective intelligence interface,\footnote{\url{https://ai4ci.ac.uk/events/}} and launching a funding opportunity to support new collaborative research activities in this space across the UK.\footnote{\url{https://ai4ci.ac.uk/funding/}} No doubt there are many viable alternative research strategies at this same interface, and we echo Nesta's conclusion that ``the field can only evolve through more organisations experimenting with different models of AI and CI and the opportunity to deliver novel solutions to real-world challenges''.\footnote{\url{https://www.nesta.org.uk/report/future-minds-and-machines/}}

\subsection*{Acknowledgements}

This work was supported by UKRI EPSRC Grant No.\ EP/Y028392/1: \textsl{AI for Collective Intelligence (AI4CI)}.

\subsection*{Competing Interests}

The authors declare none.


\bibliographystyle{agsm}
\bibliography{ai4ci.bib}

\end{document}